\documentclass[11pt,letterpaper]{article}
\usepackage[table]{xcolor}
\usepackage{xspace}

\usepackage{acl2018}

\usepackage{times}
\usepackage{inconsolata}
\usepackage{url}
\usepackage{booktabs}
\usepackage{microtype}
\usepackage{latexsym}
\usepackage{booktabs}
\usepackage[utf8]{inputenc} 
\usepackage[T1]{fontenc}
\usepackage{graphicx}
\usepackage{amsmath}
\usepackage{comment}
\usepackage[small]{caption}
\usepackage{subcaption}
\usepackage{amssymb}
\usepackage{algorithm}
\usepackage[noend]{algpseudocode}
\usepackage{titlesec}
\usepackage{dblfloatfix} 
\usepackage{caption}
\usepackage{tipa}
\usepackage{hhline}
\usepackage{multirow}
\aclfinalcopy
\def\figref#1{Figure~\ref{fig:#1}}
\def\figlabel#1{\label{fig:#1}\label{p:#1}}

\def\tabref#1{Table~\ref{tab:#1}}
\def\tablabel#1{\label{tab:#1}\label{p:#1}}

\def\secref#1{Section~\ref{sec:#1}}
\def\seclabel#1{\label{sec:#1}\label{p:#1}}
\def\eqref#1{Eq.~\ref{eqn:#1}}
\def\eqlabel#1{\label{eqn:#1}}

\usepackage{tipa,delarray,amsmath}

\newcommand{\wordfont}[1]{{\em #1}}
\newcommand{\gloss}[1]{`#1'}
\newcommand{\software}[1]{\textsc{#1}}

\newif\ifverbose
\verbosetrue
\ifverbose
\usepackage{todonotes} 
\newcommand{\rationale}[1]{\noindent{\footnotesize \colorbox{green}{{\sc rationale}:} #1}\hfill\newline}

\else
\usepackage[disable]{todonotes}   % uncomment to disable
\newcommand{\topic}[1]{}
\newcommand{\rationale}[1]{}
\fi

\usepackage{enumitem}

\newcommand{\ryan}[2][]{\Note[#1]{Ryan}{blue!40}{#2}}
\newcommand{\thomas}[2][]{\Note[#1]{Thomas}{orange!40}{#2}}
\newcommand{\alex}[2][]{\Note[#1]{Alex}{orange!40}{#2}}

\newcommand{\Thomas}[2][]{\thomas[inline,#1]{#2}}

\long\def\eat#1{\ignorespaces}
\long\def\oldenote#1#2{\ignorespaces}
\newcommand{\modelname}{\software{Chipmunk}\xspace}
\newcommand{\cistern}{{\small \url{http://cistern.cis.lmu.de/chipmunk}\xspace}}

\newcommand{\LMS}{LMS\xspace}
\newcommand{\UMS}{UMS\xspace}

\newcommand{\MTC}{morphological tag classification\xspace}
\newcommand{\dt}[2]{#2}
\newcommand{\dtl}[3]{#2}% (#1)\textsuperscript{#3}}

\eat{

\renewcommand{\Thomas}[1]{}
\renewcommand{\thomas}[1]{}
\renewcommand{\alex}[1]{}
\renewcommand{\ryan}[1]{}

\renewcommand{\Thomas}[1]{{\huge #1}}
\renewcommand{\thomas}[1]{{\huge #1}}
\renewcommand{\alex}[1]{{\huge #1}}
\renewcommand{\ryan}[1]{{\huge #1}}

}

\renewcommand{\vec}{\boldsymbol}   % optional

\newcommand{\vs}[0]{\vec{s}}
\newcommand{\vw}[0]{\vec{w}}
\newcommand{\vl}[0]{\vec{\ell}}

\makeatletter
\newcommand*{\bigcdot}{}% Check if undefined
\DeclareRobustCommand*{\bigcdot}{%
  \mathbin{\mathpalette\bigcdot@{}}%
}
\newcommand*{\bigcdot@scalefactor}{.5}
\newcommand*{\bigcdot@widthfactor}{1.15}
\newcommand*{\bigcdot@}[2]{%
  \sbox0{$#1\vcenter{}$}% math axis
  \sbox2{$#1\cdot\m@th$}%
  \hbox to \bigcdot@widthfactor\wd2{%
    \hfil
    \raise\ht0\hbox{%
      \scalebox{\bigcdot@scalefactor}{%
        \lower\ht0\hbox{$#1\bullet\m@th$}%
      }%
    }%
    \hfil
  }%
}
\makeatother

\newcommand{\vlambda}[0]{\vec{\theta}}

\newcommand{\vg}[0]{\vec{f}}
\newcommand{\likelihood}[0]{\mathcal{L}}

\newcommand{\valpha}[0]{\vec{\alpha}}

\newcommand{\defn}[1]{\textbf{#1}}
\newcommand{\defeq}{\stackrel{\mbox{\tiny def}}{=}}

\newcounter{notecounter}

\newcommand{\enoteson}{\long\gdef\enote##1##2{{
\stepcounter{notecounter}
\large \bf
\hspace{1cm}\arabic{notecounter} $<<<$ ##1: ##2
$>>>$\hspace{1cm}}}}
\enoteson
\title{
  Labeled Morphological Segmentation with Semi-Markov Models}
\author{Ryan Cotterell$^{1,2}$\quad Thomas M{\"u}ller$^{2}$ \quad  Alexander Fraser$^{2}$ \quad   Hinrich Sch{\"u}tze$^{2}$ \\
$^{1}$Johns Hopkins University \quad $^{1,2}$LMU Munich \\
\texttt{\href{mailto:ryan.cotterell@jhu.edu}{ryan.cotterell@jhu.edu}}\quad 
\texttt{\href{mailto:muellets@cis.lmu.de}{muellets@cis.lmu.de}}
}

\begin{document}

\maketitle
\begin{abstract}
We present labeled morphological segmentation, an alternative view of morphological processing that unifies several
tasks.  
From an annotation standpoint, we additionally introduce a new hierarchy of morphotactic tagsets.
Finally, we develop \modelname, a discriminative morphological segmentation system that, contrary to previous work, explicitly models morphotactics. 
We show that \modelname yields improved performance on three tasks for all six languages: (i) morphological segmentation, (ii) stemming and (iii) \MTC. 
On morphological segmentation, our method shows absolute improvements of 2--6 points $F_1$ over the baseline.
\end{abstract}

\section{Introduction\footnote{The datasets created, an additional description of our tagsets and \modelname can be found at \cistern.}}

%% genç-leş-me-ler-in
%% genç: young
%% leş:   A-N derivation "become"   --> genç-leş: to rejuvenate (literally to become young)
%% me:   N-V derivation as "-ing"     --> genç-leş-me: rejuvenating
%% ler:    plural                              --> genç-leş-me-ler: rejuvenatings 
%% in:   genitive  marker                --> genç-leş-me-ler-in: of rejuvenatings

\begin{figure*}
  \def\arraystretch{1.0}

  \centering
%  \begin{small}
  \begin{tabular}{|llllll|}
    \hline
    \rowcolor{gray!15}\multicolumn{6}{|c|}{gençleşmelerin}\\
    \hline
    \hline
    \rowcolor{yellow!15} {\bf \UMS} & genç & leş & me & ler & in \\
    \hline
    \rowcolor{orange!15}{\bf Gloss} & young & -ate & -ion & -s & \tiny\textsc{genitive marker} \\
    \hline
    \hline
    \rowcolor{red!15}{\bf \LMS} \multirow{2}{*}{\cellcolor{red!15}} & genç & leş & me & ler & in \\
     \rowcolor{red!15}  &  \tiny  \textsc{Root:Adjectival} &
            \tiny \textsc{Suffix:Deriv:Verb} & 
            \tiny\textsc{Suffix:Deriv:Noun} & 
            \tiny\textsc{Suffix:Infl:Noun:Plural} &
            \tiny\textsc{Suffix:Infl:Noun:Genitive}\\
            \hline
            \hline
            \cellcolor{green!20} {\bf Root} &
            \multicolumn{1}{l|}{\cellcolor{green!20} genç} &
            \cellcolor{cyan!20}{\bf Stem} &
            \multicolumn{1}{l|}{\cellcolor{cyan!20} gençleşme} &
            \cellcolor{violet!20}{\bf Morphological Tag} &
            \multicolumn{1}{l|}{\cellcolor{violet!20}\tiny\textsc{Plural:Genitive}} \\
    \hline
  \end{tabular}
%\end{small}
  \caption{Examples of the tasks addressed
    %in this paper
    for the Turkish word \wordfont{gençleşmelerin} (\gloss{of the rejuvenatings}): Traditional unlabeled segmentation (\UMS), Labeled morphological segmentation (\LMS), stemming / root detection and (inflectional) morphological tag classification. The morphotactic annotations produced by \LMS allow us to solve these tasks using a single model.}
 %  \vspace{-20pt}
\figlabel{tasks}
\end{figure*}

Morphological processing is often an overlooked
problem because many well-researched languages in NLP, e.g., Chinese and English, are morphologically impoverished.
However, for languages with complex morphology, e.g., Finnish and Turkish, morphological processing is essential.
A specific form of morphological processing, morphological segmentation, has shown its utility for
machine translation \cite{dyer2008}, sentiment analysis
\cite{abdulmageed14samar}, bilingual word alignment
\cite{eyigoz2013simultaneous}, speech processing
\cite{creutz2007morph} and keyword spotting
\cite{narasimhan_morphological_2014}, \textit{inter alia}.
We advance the state of the art in supervised morphological segmentation by describing a high-performance, data-driven tool for handling complex morphology, even in low-resource settings.

In this work, we make the distinction between \emph{unlabeled} morphological segmentation (\UMS), which is often just called morphological segmentation, and \emph{labeled} morphological segmentation (\LMS).  
The labels in our supervised discriminative model for \LMS capture the distinctions between different types of morphemes and directly model the morphotactics of the language. 
We further create a hierarchical universal tagset for labeling morphemes, with different levels appropriate for different tasks.
Our hierarchical tagset was designed by creating a standard representation from heterogeneous resources for six languages.
We give an overview of the tasks addressed in this paper in \figref{tasks}, which shows the expected output for the Turkish word \wordfont{gençleşmelerin} (\gloss{of the rejuvenatings}). 
In particular, it shows the full labeled morphological segmentation, from which three representations can be directly derived: the unlabeled
morphological segmentation, the stem (or root)\footnote{Terminological notes: We use \defn{root} to refer to a morpheme with concrete meaning, \defn{stem} to refer to the concatenation of all roots and derivational affixes, \defn{root detection} to refer to stripping both derivational \defn{and}
inflectional affixes, and \defn{stemming} to refer to stripping only inflectional affixes.} and the morphological tag containing part-of-speech (POS) and inflectional features.

We model these tasks with \modelname, a semi-Markov conditional random field \citep[semi-CRF;][]{cohen_semi-markov_2004}, a model that is well-suited for
morphological segmentation. 
We provide an evaluation and analysis on six languages; 
\modelname yields strong results on all three tasks, including state-of-the-art accuracy on morphological segmentation.\looseness=-1

\paragraph{Paper Outline.}
\secref{approach}
presents our \LMS framework and the
morphotactic tagsets we develop, i.e., the labels of the sequence prediction task \modelname solves.  \secref{model}  introduces our semi-CRF model.  \secref{features} presents our novel features. \secref{relatedwork} compares \modelname  to previous work. \looseness=-1
\secref{experiments} presents experiments on the three complementary tasks of segmentation (\UMS),  stemming, and \MTC. 
\secref{fsmorph} briefly discusses finite-state morphology.

\section{Labeled Segmentation and Tagset}\seclabel{approach}
We define the framework of labeled morphological segmentation, an enhancement of morphological segmentation that---in addition to identifying the boundaries of segments---assigns a fine-grained morphotactic tag to each segment.
\LMS leads to both better modeling of segmentation and subsumes several other tasks, e.g., stemming. 

Most previous approaches to morphological segmentation are either 
unlabeled or use a small, coarse-grained set such as $\{\textsc{prefix}, \textsc{root}, \textsc{suffix}\}$. 
In contrast, our labels are fine-grained.
This finer
granularity has two advantages. (i) The labels are needed for many
tasks, for instance in sentiment analysis detecting morphologically
encoded negation, as in Turkish, is crucial. 
In other words, for many applications \UMS\ is insufficient. (ii) The \LMS framework allows us to learn a probabilistic model of morphotactics. 
Working with \LMS results in higher \UMS accuracy. 
Thus, in applications that only need segments and no labels, \LMS is beneficial.
Note that the concatenation of labels {\em across} segments yields
a bundle of morphological attributes similar to those found in the CoNLL
datasets 
often used to train morphological taggers 
\cite{buchholz2006conll}---thus \LMS helps to unify \UMS and morphological tagging.
We believe that \LMS is a needed extension of current work in
morphological segmentation. Our framework 
concisely allows the model to capture
interdependencies among various morphemes and model relations between entire morpheme classes---a neglected
aspect of the problem.
%%%% AF moved to previous work
%% Note that \newcite{jappinen1986associative} describes a rule-based
%% framework that encompasses some key ideas in \LMS, but the framework was
%% only applied to Finnish and was never applied within the context of a
%% statistical model.

\begin{figure*}
\begin{small}
  \begin{subfigure}[b]{0.5\textwidth}

\begin{tabular}{rllll} \toprule

\centering
\hspace{1.5cm} 5 & \textsc{Prefix:Deriv:Verb} & \textsc{Root:Noun} & \textsc{Suffix:Deriv:Noun} & \textsc{Suffix:Infl:Noun:Plural}\\ \midrule
4 & \textsc{Prefix:Deriv:Verb} & \textsc{Root:Noun} & \textsc{Suffix:Deriv:Noun} & \textsc{Suffix:Infl:Noun:Number} \\ \midrule
3 & \textsc{Prefix:Deriv:Verb} & \textsc{Root:Noun} & \textsc{Suffix:Deriv:Noun} & \textsc{Suffix:Infl:Noun} \\\midrule
2 & \textsc{Prefix:Deriv} & \textsc{Root} & \textsc{Suffix:Deriv} & \textsc{Suffix:Infl} \\ \midrule
1 & \textsc{Prefix} & \textsc{Root} & \textsc{Suffix} & \textsc{Suffix} \\ \midrule
0 & \textsc{Segment} & \textsc{Segment} & \textsc{Segment} & \textsc{Segment} \\ \midrule
German  & \wordfont{Ent} & \wordfont{eis} & \wordfont{ung} & \wordfont{en}  \\\midrule
English  & \wordfont{de} & \wordfont{frost} & \wordfont{ing} & \wordfont{s}  \\ \bottomrule
\end{tabular} 
\end{subfigure}
\caption{Example of the different morphotactic tagset granularities
for German \wordfont{Enteisungen} \gloss{defrostings}.}
\figlabel{tags}
\vspace{-10pt}
\end{small}

\end{figure*}
%To solve the problem, w
We first create a hierarchical tagset with increasing granularity, which we created by analyzing heterogeneous resources for the six languages we work on.
The optimal level of granularity is
task- and language-dependent: the level is a trade-off between
simplicity and expressivity. 
We illustrate our tagset with the decomposition of
the German word \wordfont{Ent\-eisungen} \gloss{defrostings} 
(\figref{tags}): 

\begin{itemize}[leftmargin=*]
\item \textbf{Level 0}: The level 0 tagset involves a single tag indicating a segment. 
It ignores morphotactics completely and is
similar to previous work.\looseness=-1
\item  \textbf{Level 1}: The level 1 tagset crudely approximates
morphotactics: it consists of the tags
$\{$\textsc{Prefix}, \textsc{Root}, \textsc{Suffix}$\}$.  
This scheme
has been successfully used by unsupervised segmenters, e.g.,
\software{Morfessor CAT-Map} \cite{creutz_unsupervised_2007}. 
It allows the model to learn simple morphotactics, for instance that a
prefix cannot be followed by a suffix. This makes a decomposition like
\wordfont{reed} $\mapsto$ \wordfont{re}+\wordfont{ed} unlikely.
We also add an additional \textsc{Unknown} tag for morphemes that do
not fit into this scheme.\looseness=-1
\item  \textbf{Level 2}: The level 2  tagset splits
affixes into \textsc{Derivational} and \textsc{Inflectional},
effectively increasing the maximal tagset size from 4 to 6. 
These tags can encode that many languages allow for transitions from
derivational to inflectional endings, but rarely the opposite. 
This
makes the incorrect decomposition of German 
\wordfont{Offenheit}
(\gloss{openness}) into \wordfont{Off}, inflectional \wordfont{en} and
derivational \wordfont{heit} unlikely.\footnote{In \wordfont{open} (English) and \wordfont{Offen} (German), the \wordfont{en} is part of the root.\looseness=-1}  
This tagset is also useful for building statistical stemmers.  
\item  \textbf{Level 3}: The level 3 tagset adds the part of speech, i.e., whether a root is \textsc{Verbal}, \textsc{Nominal} or
\textsc{Adjectival}, and the part of speech of the word that an affix
derives. 
\item  \textbf{Level 4}: The level 4 tagset
includes the inflectional feature a suffix adds, e.g.,
\textsc{Case} or \textsc{Number}. 
This is helpful for certain agglutinative
languages, in which, e.g.,
\textsc{Case} must follow \textsc{Number}.\looseness=-1
\item  \textbf{Level 5}: The level 5 tagset adds the actual value of
the inflectional feature, e.g., \textsc{Plural},
and corresponds to the annotation in the datasets.
In preliminary experiments we found that the level 5 tagset
is too rich and does not yield consistent improvements; we
thus do not report experimental results using it.
\end{itemize}
\tabref{tagsetsizes} shows tagset sizes for the six languages.\footnote{As converting segmentation datasets to tagsets is not always straightforward, we include tags that lack some features, e.g., some level 4 German tags lack POS because our German data does not specify it.\looseness=-1
}

% Table generated by:
% $ for lang in eng fin ger ind tur zul; do for i in 0 1 2 3; do echo -n "$lang $i "; cut -f3 /mounts/data/proj/marmot/ryan/segmentation/data_dev/$lang/0.trn.newtags.$i.feats  | sed -e 's/^[BIS]|//g' | sed -e s/_/UNK/g | grep -ve '^$' | uniq | sort | uniq | wc -l; done; done

%% \Jason{consider transposing \tabref{tagsetsizes} so that the rows are numbered 5,4,\ldots 0
%%   as in \figref{tags}.  Actually, what happened to granularity 5 in
%%   this table??}

%% \enote{hs}{i'm not sure how much we want to emphasize this
%% aspect -- if additional resources are available for  a
%% language, presumably we want to use them. conversely, lots
%% of languages are not covered, e.g.,  Swiss German and Guarani
%% Our features are based on linguistic resources that are
%% either freely available for most languages or can be quickly
%% created. The morphemes of a language can usually be
%% subdivided into open-class set of stems and a closed-class
%% set of productive derivational and inflectional affixes.
%% We cover the closed set of affixes by
%% extraction gazetteers lists from Wiktionary.
%% }

\section{Model}
\seclabel{model}
%\enote{Helmut}{Is ``Cemino'' an acronym?}
%We present
\modelname is a supervised model based on a
semi-Markov conditional random field (semi-CRF)
\cite{cohen_semi-markov_2004} that naturally fits
the task of \LMS.
%our framework. 
%\modelname\ outperforms prior work on unlabeled morphological
%segmentation. 
%\subsection{Semi-CRFs}
%% \enote{hs}{can't CRFs also model segmentation? i was
%%   surprised by this intro and seems to contradict the first
%%   paragraph of ``approach''}
Semi-CRFs generalize linear-chain CRFs and model segmentation jointly with sequence labeling.  
Just as linear-chain CRFs are discriminative adaptations of \defn{hidden Markov models} \cite{lafferty_conditional_2001}, semi-CRFs are an analogous adaptation of \defn{hidden semi-Markov models} \cite{murphy2002hidden}. 
Semi-CRFs allow us to integrate new features that look at complete segments, which is not possible with CRFs, making semi-CRFs a natural choice for morphology.\looseness=-1

A semi-CRF represents $\vw$ (a word) as a sequence of segments $\vs =
\langle s_1,\ldots,s_N \rangle$, each of which is assigned a label $\ell_n$.
The concatenation of all segments equals $\vw$.  
We seek a log-linear distribution $p_{\vlambda}(\vs, \vl \mid \vw)$ over all possible segmentations and label sequences for
$\vw$, where $\vlambda$ is the parameter vector. 
Note that we recover the standard  CRF if we restrict the segment length to 1. 
Formally, we define the probability distribution $p_{\vlambda}$ as
\begin{align}
p_{\vlambda}(\vs,\vl \mid \vw) \defeq \frac{1}{Z_{\vlambda}(\vw)}\prod_{n=1}^N \exp\left(
\vlambda \bigcdot \vg_n\right), 
\end{align}
where $\vg_n \defeq \vg(s_n,\ell_n,\ell_{n-1},n)$ is the feature function and $Z_{\vlambda}(\vw)$ is the partition function. 
We use a generalization of the forward-backward algorithm
for efficient gradient computation \cite{cohen_semi-markov_2004}.
Inspection of the semi-Markov forward recursion, 
\begin{align}\eqlabel{recursion}
\valpha(0,\ell) &= 1 \qquad \qquad \qquad \qquad{\color{gray}(\forall \ell \in [L])} \\
 \valpha(n,\ell) &= \sum_{t=1}^{n} \sum_{\ell'=1}^L \exp (\vlambda \bigcdot  \vg_n) \cdot \valpha(n-t,\ell'), \nonumber
\end{align}
 shows that algorithm runs in $\mathcal{O}(N^2  L^2)$ time where $N$ is the
 length of the word $\vw$ and $L$ is the number of labels (size of the tagset). 
 Then, we have the partition function equals $Z_{\vlambda}(\vw) = \sum_{\ell=1}^L \valpha(N, \ell)$.
A similar recursion, generalizing the Viterbi algorithm for hidden Markov models \citep{rabiner1989tutorial}, allows us to find the one-best labeled segmentation in $\mathcal{O}(N^2  L^2)$ as well.\looseness=-1

We employ the maximum-likelihood criterion to 
estimate the parameters with L-BFGS \cite{liu1989limited},
a gradient-based optimization algorithm.
As in all exponential family models,
the gradient of the log-likelihood takes the form of the difference
between the observed and expected feature counts \cite{wainwright2008graphical}
and can be computed efficiently with the semi-Markov
extension of the forward-backward algorithm.
%\alex{this formula looks ugly, left justify?}
%\begin{equation}
%\frac{\partial}{\partial \vlambda} \likelihood(\vlambda) = \sum_i  \vg - \nonumber \\   \sum_{\vs',\vl'} \sum_i  p_{\vl%ambda}(s'_i,\ell'_i \mid \ell'_{i-1},\vw)\cdot \vg.  \nonumber
% \end{equation} 
We use $L_2$ regularization with a regularization coefficient tuned during
cross-validation. 

We note that semi-Markov models have the potential to obviate typical
errors made by standard Markovian sequence models with an IOB labeling
scheme
%.  
over characters.
For instance, consider the incorrect segmentation of the
English verb \wordfont{sees} into \wordfont{se+es}. These are
reasonable split positions as many English stems end in \wordfont{se}
(e.g., consider \wordfont{abuse+s}).  
Semi-CRFs have a major advantage because they admit \emph{segmental} features that allow them to learn \wordfont{se} is not a good morph.\looseness=-1

\begin{table}[t]
\begin{center}
\begin{small}
\begin{tabular}{llrrrrrr} \toprule
           & & \multicolumn{5}{c}{tagset level}\\
 & language        & 0 & 1 & 2 & 3 & 4 \\
\midrule
&English     & 1 & 4 & 5 & 13 & 16\\
&Finnish     & 1 & 4 & 6 & 14 & 17\\
&German      & 1 & 4 & 6 & 13 & 17\\
 &Indonesian  & 1 & 4 & 4 &  8 &  8\\
&Turkish     & 1 & 3 & 4 & 10 & 20\\ 
&Zulu        & 1 & 4 & 6 & 14 & 17 \\ \bottomrule
\end{tabular}
\end{small}
\end{center}
\caption{Morphotactic tagset size at each level of granularity.}
\tablabel{tagsetsizes}
\end{table}

\section{Features}
\seclabel{features}
We introduce several novel features
for \LMS. We exploit existing 
resources, e.g., spell checkers and Wiktionary, to create straightforward
and effective features and 
we
incorporate ideas
from related areas: named-entity recognition (NER) and morphological tagging.\looseness=-1

\paragraph{Affix Features and Gazetteers.}
In contrast to syntax and semantics, the morphology of a language is
often 
%quite 
simple to document and a list of the most common morphs
can be found in any good grammar. 
Wiktionary, for example,
contains affix lists for all the six languages used in our
experiments.\footnote{A good example of such a
    resource is \url{en.wiktio-} \url{nary.org/wiki/Category:Turkish_suffixes}.}
Providing a supervised learner with
such a list is a great boon, just as gazetteer features aid
NER \cite{smith_using_2006}.
The benefit is perhaps even greater than in applications
like NER because suffixes and prefixes are generally closed-class, and hence
these lists are likely to be comprehensive.  
These features are binary and fire if a given substring occurs in the gazetteer list.
In this paper, we simply use suffix lists from  English
Wiktionary, except for Zulu, for which we use a prefix list, see \tabref{gazetteers}.
We also include a feature that fires on the conjunction
of tags and substrings observed in the training data. In the level 5 tagset, this allows us to link all allomorphs of a given morpheme. 
In the lower-level tagsets, this links related morphemes.
\newcite{virpioja2010unsupervised} explored this idea for unsupervised
segmentation. Linking allomorphs together under a single tag
helps combat sparsity in modeling the morphotactics.

%% \enote{hs}{i find ``a binary feature'' not very clear
%% i guess you introduce one binary feature for each tag
%% occurring in the training set?}
%% \jason{i had the same question as Hinrich}

\begin{table}
\begin{small}
\begin{tabular}{lrl} \toprule
        & \#  affixes & random examples\\
\midrule
English    & 394 & -\wordfont{ard} -\wordfont{taxy} -\wordfont{odon} -\wordfont{en} -\wordfont{otic} -\wordfont{fold} \\
Finnish    & 120 & -\wordfont{tä} -\wordfont{llä} -\wordfont{ja} -\wordfont{t} -\wordfont{nen} -\wordfont{hön} -\wordfont{jä} -\wordfont{ton}\\
German     & 112 & -\wordfont{nomie} -\wordfont{lichenes} -\wordfont{ell} -\wordfont{en} -\wordfont{yl} -\wordfont{iv}\\
Indonesian &   5 & -\wordfont{kau} -\wordfont{an} -\wordfont{nya} -\wordfont{ku} -\wordfont{mu}\\
Turkish    & 263 & -\wordfont{ten} -\wordfont{suz} -\wordfont{mek} -\wordfont{den} -\wordfont{t} -\wordfont{ünüz}\\
Zulu       &  72 & \wordfont{i}- \wordfont{u}- \wordfont{za}- \wordfont{tsh}- \wordfont{mi}- \wordfont{obu}- \wordfont{olu}- \\ \bottomrule
\end{tabular}
\end{small}
\caption{
%Frequencies 
Sizes
of the various affix gazetteers.}\vspace{-17.5pt}

\tablabel{gazetteers}
\end{table}

\paragraph{Stem Features.}
A major problem in statistical segmentation is the reluctance to
posit morphs not observed in training; this particularly
affects roots, which are {\em open-class}. %% \enote{hs}{i cut this
%%   because it's true for compounding in general? For languages with heavy
%% compounding, e.g., English and German,} \jason{however, note that these features won't
%% successfully recognize a root that changes in context, as in Japanese
%% ori+kami -> origami, or English home+ing -> homing.}
This makes it nearly impossible
to correctly segment compounds that contain  unseen
roots, e.g., to correctly segment \wordfont{homework} you
need to know that
\wordfont{home} and \wordfont{work}
are independent English words. We solve this problem 
by incorporating spell-check features: binary features
that fire if a segment is valid for a given spell checker.
Spell-check features act as an effective proxy for a root detector. 
We use the open-source
\software{aspell} dictionaries as they are freely available in 91
languages. \tabref{aspell} shows the coverage.\looseness=-1

\paragraph{Integrating the Features.}  
Our  model uses the features discussed in this section and
additionally the simple $n$-gram context features of
\newcite{ruokolainen_supervised_2013}. The $n$-gram features look at variable length substrings
of the word on both the right and left side of each boundary.
We create conjunctive features from the cross-product between
the morphotactic tagset (\secref{approach}) and the features.

%% \jason{This is a bit imprecise.  The Cartesian product is a set of
%%   {\em pairs}; but you are building a set of {\em conjunctions}.  You also haven't said
%%   exactly how you use them in the CRF model after you ``build'' them:
%%   shouldn't you mention $\vg(\vw,s_i,\ell_i,\ell_{i-1},i)$ again?
%%   And all of the features you've listed seem to depend only on $s_i$
%%   and $\ell_i$ (by conjunction), so what are your features that depend on
%%   $\ell_{i-1}$ and/or $i$?  Finally, for readers who think that features are the things that get
%%   weights, the only ``features'' come from this Cartesian product.}
%To reduce sparsity we additionally decompose the rich tags (\figref{tags})
%into their subtags and combine these with the subtags of the previous
%tag.  This is similar to work on POS tagging
%for highly-inflected languages
%\cite{muller_efficient_2013,silfverberg2014part}.

\section{Related Work}
\seclabel{relatedwork}

% AF moved this to top with other LMS works
%\newcite{jappinen1986associative} describe a rule-based
%framework that encompasses some key ideas in \LMS, but the framework was
%only applied to Finnish and was never applied within the context of a
%statistical model.

%Most similar to this work are 

\paragraph{Memory-based Learning.}
\newcite{van1999memory} and \newcite{marsi2005}
present memory-based approaches to discriminative learning of morphological segmentation and both address the problem of \LMS.
We distinguish our work from theirs in that we define a cross-lingual schema for defining a hierarchical tagset for \LMS. 
Morever, we tackle the problem with a feature-rich, log-linear  model, allowing us to easily incorporate disparate sources of knowledge into a single framework.\looseness=-1

\begin{table}
  \centering
  \begin{small}
  \begin{tabular}{lr} \toprule
  language & \# words \\ \midrule
    English    &    119,839 \\
    Finnish    &  6,690,417 \\
    German     &    364,564 \\
    Indonesian &     35,269 \\
    Turkish    &     80,261 \\
    Zulu       &     73,525 \\ \bottomrule
  \end{tabular}
  \end{small}
  \caption{Number of words covered by the \software{aspell} dictionary}
  \tablabel{aspell}
 \vspace{-15pt}
\end{table}

\paragraph{Unsupervised \UMS.}
\UMS has been mainly addressed by unsupervised algorithms.
\software{Linguistica} \cite{goldsmith_unsupervised_2001}
and \software{Morfessor} \cite{creutz2002unsupervised} are
built around the idea of optimally encoding the data, in the
sense of minimal description length (MDL).
\software{Morfessor Cat-MAP} \cite{creutz_unsupervised_2007} formulates the model as sequence prediction based on HMMs over a morph dictionary and MAP estimation.  
The model also attempts to induce basic morphotactic categories (\textsc{Prefix}, \textsc{Root}, \textsc{Suffix}).  
\citet{kohonen_semi-supervised_2010,kohonen2010} and \newcite{gronroos_morfessor_2014} present variations of \software{Morfessor} for semi-supervised learning.
\newcite{poon_unsupervised_2009} introduces a Bayesian state-space model
with corpus-wide priors. The model resembles a semi-CRF, but 
dynamic programming is no longer tractable. 
They employ the three-state tagset of \newcite{creutz2004} (row 1 of \figref{tags}) for Arabic and Hebrew \UMS. 
Their gradient and objective computation is based on an enumeration of a heuristically chosen subset of the exponentially many segmentations.
This limits its applicability to language with complex \emph{concatenative} morphology, e.g., Turkish and Finnish.\looseness=-1

\paragraph{Supervised \UMS.}
\newcite{ruokolainen_supervised_2013} present an
averaged perceptron \cite{collins2002discriminative}, a discriminative structured prediction method, for \UMS.
The model outperforms the semi-supervised model of \newcite{poon_unsupervised_2009} on Arabic and Hebrew morpheme segmentation as well as the semi-supervised model of \newcite{kohonen2010} on English, Finnish and Turkish. 
\newcite{ruokolainen_painless_2014} get further
empirical improvements by using features extracted from large corpora, based on the letter successor variety (LSV) model \cite{harris1955} and on unsupervised segmentation models such as Morfessor CatMAP \cite{creutz_unsupervised_2007}.
The idea behind LSV is that for example \wordfont{talking} should be split into \wordfont{talk} and \wordfont{ing}, because \wordfont{talk} can also be followed by different letters then \wordfont{i} such as \wordfont{e} (talked) and \wordfont{s} (talks).\looseness=-1

\paragraph{Chinese Word Segmentation.}
Chinese word segmentation
(CWS) is related to \UMS.
\newcite{andrew_cws_2006} successfully apply semi-CRFs to
CWS. 
Joint CWS and POS tagging \cite{ng2004chinese,zhang2008joint} is
related to \LMS.\looseness=-1
%To our knowledge, joint CWS and POS tagging
%has not been modeled with a semi-CRF.

\begin{table}
\begin{small}
\begin{center}
\begin{tabular}{lrrrrr} \toprule
     %      & & \multicolumn{3}{c}{Train+Tune+Dev} & Test \\
%\midrule
           &         un. data        & train & tune & dev     &    test  \\  
\midrule
English    &        878k & 800  &   100 &  100      &  694 \\
Finnish    &      2,928k & 800  &   100 &  100      &  835 \\
German     &      2,338k & 800  &   100 &  100      &  751 \\
Indonesian  &         88k & 800  &   100 &  100      & 2500 \\
Turkish    &        617k & 800  &   100 &  100      &  763 \\
Zulu       &        123k & 800  &   100 &  100      & 9040 \\  \bottomrule
\end{tabular}
\end{center}
\caption{Dataset sizes (number of types).}
\tablabel{datasetsizes}
\end{small}
\vspace{-19pt}

\end{table}
\begin{table}
\begin{small}
\begin{center}
\begin{tabular}{rrrc} \toprule
& & +Affix & +Dict,+Affix   \\
\midrule
Level 0    &        90.11   & 90.13   &   91.66   \\
Level 1    &      90.73    & 90.68   &   92.80  \\
Level 2     &      89.80   & 90.46   &   92.04  \\
Level 3  &         91.03   & 90.83   &   92.31  \\
Level 4    &        91.80  & 92.19   &   \textbf{93.21} \\ \bottomrule
\end{tabular}
\end{center}
\caption{Example of the effect of larger tagsets (\figref{tags}) on Turkish segmentation measured on our development set. 
As Turkish is an agglutinative language with hundreds of affixes, the efficacy of our approach is expected to be particularly salient here. 
Recall we optimized for the best tagset granularity for our experiments on Tune.}
\vspace{-10pt}
\tablabel{turkishcurves}
\end{small}
\end{table}

\begin{table*}
\centering
\begin{small}
\begin{tabular}{  l  c  c  c  c  c  c }  \toprule
                                & English           & Finnish           & Indonesian       & German            & Turkish           & Zulu \\ \midrule
\software{CRF-Morph}            & \dt{86.10}{83.23} & \dt{82.09}{81.98} &\dt{93.01}{93.09} & \dt{85.67}{84.94} & \dt{88.05}{88.32} & \dt{86.48}{88.48} \\ 
\software{CRF-Morph} +LSV       & \dt{86.71}{84.45} & \dt{84.09}{84.35} &\dt{93.32}{93.50} & \dt{86.73}{86.90} & \dt{89.47}{89.98} & \dt{87.99}{89.06} \\ 
First-order CRF                 & \dt{86.45}{84.66} & \dt{82.53}{85.05} &\dt{93.57}{93.31} & \dt{86.16}{85.47} & \dt{87.50}{90.03} & \dt{87.32}{88.99} \\
Higher-order CRF                & \dt{86.45}{84.66} & \dt{82.34}{84.78} &\dt{93.49}{93.88} & \dt{86.90}{85.40} & \dt{88.88}{90.65} & \dt{87.50}{88.85} \\ \midrule
\modelname                     & \dt{86.34}{84.40} & \dt{82.34}{84.40} &\dt{93.55}{93.76} & \dt{84.63}{85.53} & \dt{90.11}{89.72} & \dt{87.39}{87.80} \\ 
\modelname +Morph              & \dtl{86.25}{83.27}{2} & \dtl{83.11}{84.71}{4} &\dtl{93.83}{93.17}{2} & \dtl{84.78}{84.84}{2} & \dtl{91.82}{90.48}{4} & \dtl{89.76}{90.03}{4} \\ 
\modelname +Affix              & \dt{86.15}{83.81} & \dt{83.98}{86.02} &\dt{93.94}{93.51} & \dt{85.07}{85.81} & \dt{90.73}{89.72} & \dt{88.82}{89.64} \\ 
\modelname +Dict               & \dt{87.59}{86.10} & \dt{85.38}{86.11} &\dt{95.17}{95.39} & \dt{88.27}{87.76} & \dt{90.49}{90.45} & \dt{87.35}{88.66} \\ 
\modelname +Dict,+Affix,+Morph & \dtl{{\bf 88.71}}{\bf 86.31}{2} & \dtl{{\bf 86.92}}{{\bf 88.38}}{4} &\dtl{{\bf 95.70}}{\bf 95.41}{2} & \dtl{{\bf 89.75}}{{\bf 87.85}}{2} & \dtl{{\bf 93.21}}{{\bf 91.36}}{4} & \dtl{{\bf 90.11}}{{\bf 90.16}}{4} \\  \bottomrule
\end{tabular}
\end{small}
%\end{small}
\caption{\dt{Dev}{Test} $F_1$ for \UMS. Features: LSV =
letter successor variety, Affix = affix, Dict = dictionary,
Morph = optimal (on Tune) morphotactic tagset.}
\tablabel{seg-test}
\vspace{-10pt}
\end{table*}

\section{Experiments}
\seclabel{experiments}
We experiment on six languages from diverse 
language families.
The segmentation data for English, Finnish and  Turkish was taken from MorphoChallenge 2010 \cite{kurimo2010}.\footnote{\url{http://research.ics.aalto.fi/events/morphochallenge2010/}}
Despite typically being used for \UMS tasks, 
the MorphoChallenge datasets \emph{do} contain morpheme-level labels. 
The German data was extracted from the CELEX2 collection \cite{baayen1993celex}, which contains all the requisite information.
The Zulu data was taken from the Ukwabelana 
corpus \cite{spiegler2010ukwabelana}. 
Finally, the Indonesian portion was created by applying the rule-based analyzer \software{MorphInd} \cite{larasati2011indonesian}
to the Indonesian portion of an Indonesian--English bilingual corpus.\footnote{\url{https://github.com/desmond86/Indonesian-English-Bilingual-Corpus}}

We did not have access to the MorphoChallenge test set, and, thus, we used the original development set as our final evaluation set (Test).
We developed \modelname using 10-fold cross-validation on the 1000-word
training set and split every fold into training (Train), tuning (Tune)
and development sets (Dev).\footnote{We used both Tune and Dev in order to both optimize hyperparameters on held-out data (Tune) and perform qualitative error analysis on separate held-out data (Dev).} 
For German, Indonesian and Zulu, we randomly selected 1000 word forms as training set and used the rest as
evaluation set.  
For our final evaluation we trained \modelname on
the concatenation of Train, Tune and Dev (the original 1000 word
training set), using the optimal parameters from the cross-evaluation
and tested on Test. 
\tabref{datasetsizes} shows the important statistics of
our datasets.
One of our baselines also uses unlabeled training
data. 
MorphoChallenge provides word lists for English, Finnish, German
and Turkish.  
We use the unannotated part of Ukwabelana for Zulu; and
for Indonesian, data from Wikipedia and the corpus of \newcite{krisnawati2013}.\looseness=-1

In all evaluations, we use variants of the standard MorphoChallenge
evaluation approach. 
Importantly, for word types with multiple correct
segmentations, this involves finding the maximum score by comparing
the one-best segmentation under \modelname, as computed by the Viterbi algorithm, with \emph{each} correct segmentation, as is
standardly done in MorphoChallenge.

\subsection{\UMS Experiments}
We first evaluate \modelname on \UMS, by predicting \LMS and then discarding the labels.
Our primary baseline is the state-of-the-art supervised system \software{CRF-Morph} of \newcite{ruokolainen_supervised_2013}.
We ran the version of the system that the authors published on their
website.\footnote{\url{http://users.ics.tkk.fi/tpruokol/software/crfs_morph.zip}}
We optimized the model's
two hyperparameters on Tune: the
number of epochs and the maximal length of $n$-gram character
features. 
The system also supports \citeposs{harris1955} letter successor variety (LSV) features, extracted from large unannotated corpora.
For completeness, we also compare \modelname with
a first-order CRF and a
higher-order CRF \cite{muller_efficient_2013}, both
used the same $n$-gram features as \software{CRF-Morph}, but without
the LSV features.\footnote{Model order, maximal
character $n$-gram length and regularization coefficients
were optimized on Tune.}
We evaluate all models using the traditional macro $F_1$ of the segmentation boundaries.\looseness=-1

\paragraph{General Discussion.}
The \UMS results on held-out data are displayed in
\tabref{seg-test}.  Our most complex model beats the best baseline by
between 1 (German) and 3 (Finnish) points $F_1$ on all six languages. We additionally
provide extensive ablation studies to highlight the contribution of
our novel features. We find that the properties of each specific
language highly influences which features are most effective. For the
agglutinative languages, i.e, Finnish, Turkish and Zulu, the affix-based features (+Affix) and the morphotactic tagset (+Morph) yield consistent
improvements over the semi-CRF models with a single state.
Improvements for the affix features range from 0.2 for Turkish to 2.14 for Zulu.
The morphological tagset yields improvements of 0.77 for Finnish, 1.89 for Turkish and
2.10 for Zulu. We optimized tagset granularity 
on Tune and found that levels 4 and level 2 yielded the best results
for the three agglutinative and the three other
languages, respectively.
The dictionary features (+Dict) help universally, but their effects are
particularly salient in languages with productive compounding, i.e.,
English, Finnish and German, where we see improvements of $>1.7$.
In comparison with previous work \cite{ruokolainen_supervised_2013}, we find that
our most complex model yields consistent improvements over \software{CRF-Morph} +LSV for all languages: The improvements range from $>1$ for German over $>1.5$ for Zulu, English, and Indonesian to $>2$ for Turkish and $>4$ for Finnish.\looseness=-1

\paragraph{The Role of Morphotactics.}
To illustrate the effect of modeling morphotactics through
the larger morphotactic tagset on performance, we provide a detailed analysis of Turkish.  
See \tabref{turkishcurves}. 
We consider three different feature sets and increase the size of the
morphotactic tagsets depicted in \figref{tags}.
The results evince the general trend that improved morphotactic modeling
benefits segmentation. 
Additionally, we observe that the improvements are complementary to those from the other features.\looseness=-1

\paragraph{Novel Roots and Affixes.}
As discussed earlier, a key problem in \UMS,
especially in low-resource settings, is the detection of novel
roots and affixes. 
Since many of our features were designed to combat
this problem specifically, we investigated this aspect
independently. \tabref{rooting} shows the number of
novel roots and affixes found by our best model and the baseline.
In all languages, \modelname correctly identifies between
5\% (English) and 22\% (Finnish) more novel roots than the
baseline. 
We do not see major improvements for affixes, 
but this is of less interest as there are far fewer novel affixes.

\paragraph{Boundaries.}
We further explore how \modelname and the baseline perform on different
boundary types by looking at missing boundaries between
different morphotactic types;  this error type is also known as {\em
undersegmentation}. \figref{heatmap_under} shows a heatmap that
overviews errors broken down by morphotactic tag. We see that most errors are
caused between root and suffixes across all languages. This is related
to the problem of finding new roots, as a new root is often mistaken
as a root-affix composition.\looseness=-1

\begin{figure}
\hspace{-0.25cm}
\includegraphics[width=9cm]{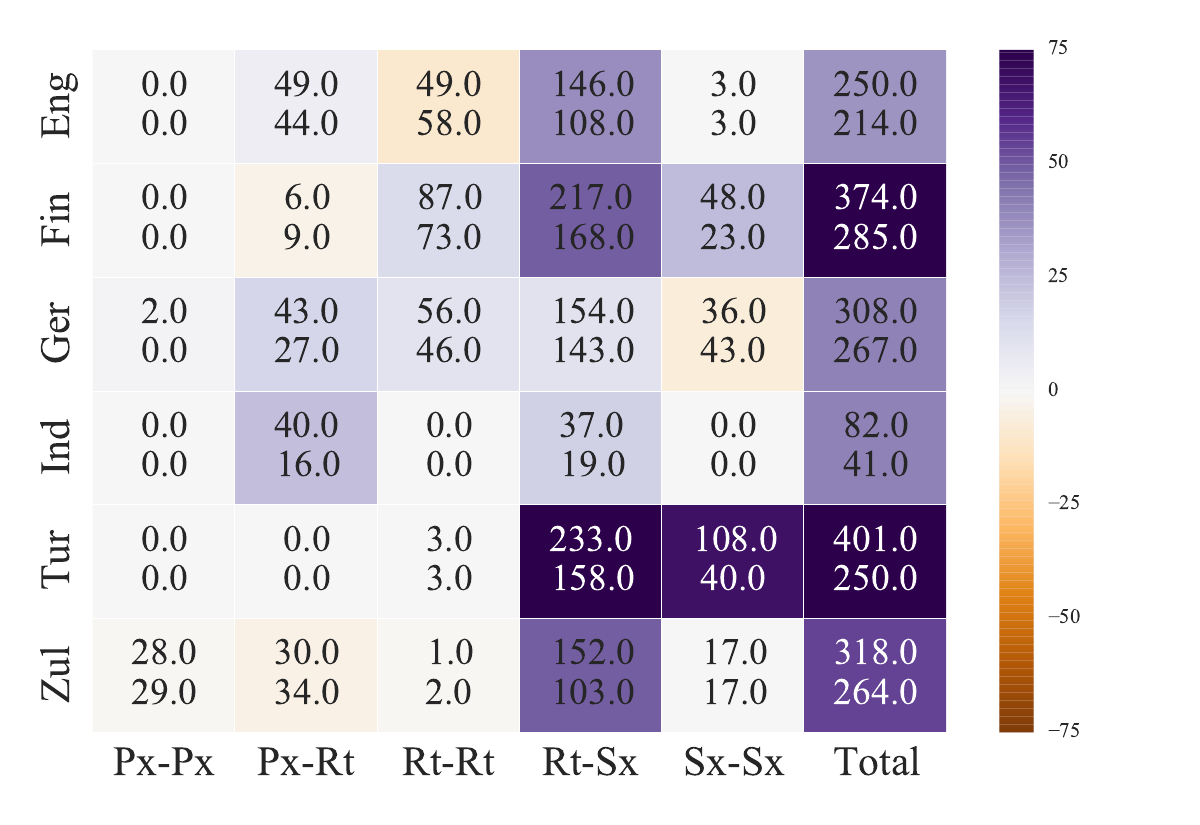}
\caption{This figure represents a comparative analysis of
undersegmentation. Each column (labels at the bottom) shows how
often \software{CRF-Morph} +LSV (top number in heatmap) and \modelname (bottom number in heatmap) select a segment that is two separate segments in the gold
standard. 
E.g., Rt-Sx indicates how a root and a suffix were treated as a single segment. The color depends on the difference of the two counts.\looseness=-1
}
\figlabel{heatmap_under}
\vspace{-15pt}
\end{figure}

\subsection{Root Detection and Stemming}
\seclabel{rootstem}

Root detection${}^1$ and stemming${}^1$ are two important
NLP problems that are
closely related to morphological segmentation and used in applications
such as MT, information retrieval, parsing and information extraction.
Here we explore the utility of 
\modelname as a statistical stemmer and root detector.
Stemming is closely related to the task of {\em lemmatization},
which involves the additional step of normalizing to the canonical
form.\footnote{In our experiments there are {\em no} stem alternations.
The output is equivalent to that of the Porter stemmer \cite{porter1980algorithm}.\looseness=-1}  
Consider the German particle verb participle
\wordfont{auf-ge-schrieb-en} \gloss{written down}. The participle is
built by applying an alternation to the verbal root \wordfont{schreib}
\gloss{write} adding the participial circumfix \wordfont{ge-en} and
finally adding the verb particle \wordfont{auf}. In our
segmentation-based definition, we would consider \wordfont{schrieb}
\gloss{write} as its root and \wordfont{auf-schrieb} as its stem. In
order to additionally to restore the lemma, %\jason{clarify: which we don't do}
we would also have to reverse the stem
alternation that replaced \wordfont{ei} with \wordfont{ie} and add the
infinitival ending \wordfont{en} yielding the infinitive
\wordfont{auf-schreib-en}.

%\begin{table}[H]
\begin{table}
\begin{small}
\centering
\begin{tabular}{lrrrr} \toprule
           & \multicolumn{2}{c}{\software{CRF-Morph}} & \multicolumn{2}{c}{\modelname} \\
           & Roots & Affixes    & Roots    & Affixes\\
\midrule
English    & 614 &  6 & 644 &  12\\
Finnish    & 502 & 10 & 613 &  11\\
German     & 360 &  6 & 414 &   9\\
Indonesian & 593 &  0 & 639 &   0\\
Turkish    & 435 & 22 & 514 &  19\\
Zulu       & 146 & 10 & 160 &  11\\ \bottomrule
\end{tabular}
\caption{Dev number of unseen root and affix types correctly identified by \software{CRF-Morph} +LSV and \modelname +Affix,+Dict,+Morph.}
\tablabel{rooting}
\end{small}
\vspace{-17.5pt}
\end{table}

%\caption{Dev number of unseen root and affix types correctly
%  identified by \software{CRF-Morph} +LSV and
%  \modelname +Affix,+Dict,+Morph.}

Our baseline \software{morfette} \cite{chrupala2008} is a statistical
transducer that first extracts edit paths between input and
output and then uses a perceptron classifier to decide which edit path
to apply. In short, \software{morfette} treats the task as a
string-to-string transduction problem, whereas we view it as a labeled
segmentation problem.\footnote{Note that \software{Morfette} is a pipeline
that first tags and {\em then} lemmatizes. We only make use of this second
part of \software{Morfette} for which it is a strong string-to-string transduction baseline.\looseness=-1
}
Note that \software{morfette} would in principle be able to handle stem alternations,
although these usually lead to an increase in the number of edit paths.
We use level 2 tagsets for all experiments---the smallest
tagsets complex enough for stemming---and extract the relevant segments.\looseness=-1

%\Jason{shouldn't you use \modelname to extract \wordfont{aufschrieb} and then postprocess with \software{morfette} (for example) to convert that to \wordfont{aufschreiben}?}

\begin{table*}
\centering
\begin{small}
\begin{tabular}{lcllllll} \toprule

                   &  &  English & Finnish & German & Indonesian & Turkish & Zulu \\
\midrule
Root & \software{morfette} &  \dt{68.60}{62.82} & \dt{33.30}{39.28} & \dt{45.20}{43.81}  & \dt{86.50}{86.00}  & \dt{27.70}{26.08} & \dt{28.20}{30.76}\\
Detection & \modelname            & {\bf \dt{73.30}{70.31}} & {\bf \dt{49.30}{69.85}} &{\bf \dt{69.60}{67.37} } & {\bf \dt{90.70}{90.00}}  & {\bf \dt{83.60}{75.62} }& {\bf \dt{57.10}{62.23}}\\
\midrule
Stemming & \software{morfette} &  \dt{93.50}{91.35} & \dt{48.50}{51.74} & \dt{79.20}{79.49} & \dt{86.50}{86.00} & \dt{28.30}{28.57} & \dt{54.70}{58.12} \\
& \modelname         &  {\bf \dt{94.60}{94.24}} & {\bf \dt{56.40}{79.23}} &{\bf \dt{85.40}{85.75}} & {\bf \dt{90.70}{89.36}} & {\bf \dt{87.80}{85.06}} &{\bf  \dt{67.45}{67.64}} \\ \bottomrule
\end{tabular}
\end{small}
\caption{Accuracy on the root detection and stemming on \dt{Dev}{Test}.}
\tablabel{stemming}
\vspace{-10pt}
\end{table*}

\begin{table}
\centering
\begin{small}
\def\arraystretch{1.25}
\begin{tabular}{llcc} \toprule
 &  & Finnish & Turkish \\
\midrule
$F_1$ & MaxEnt             & \dt{70.13}{75.61} & \dt{70.34}{69.92} \\
& MaxEnt +Split      & \dt{71.10}{74.02} & \dt{76.73}{76.61} \\
\cline{2-4}
& \modelname +All      & {\bf \dt{82.47}{80.34}} & {\bf \dt{83.98}{85.07}} \\
\midrule

Acc. & MaxEnt             & \dt{53.40}{60.96} & \dt{38.10}{37.88} \\
& MaxEnt +Split      & \dt{54.40}{59.04} & \dt{43.60}{44.30} \\
\cline{2-4}
& \modelname +All      & {\bf \dt{65.10}{65.00}} & {\bf \dt{58.40}{56.06}} \\ \bottomrule
\end{tabular}
\end{small}
\caption{\dt{Dev}{Test} results on \MTC.}
\tablabel{mtc}
\vspace{-15pt}
\end{table}

\paragraph{Discussion.}
Our results are shown in \tabref{stemming}.
We see consistent improvements 
across all tasks. For the fusional languages
(English, German and Indonesian) we see modest gains in performance on
both root detection and  stemming. However, for the
agglutinative languages (Finnish, Turkish and Zulu) we see absolute
gains as high as 50\% (Turkish) in accuracy. This significant
improvement is due to the complexity of the tasks in these
languages---their productive morphology increases  sparsity and
makes the unstructured string-to-string transduction approach
suboptimal. We view this as solid evidence that labeled segmentation
has utility in many components of the NLP pipeline.

\subsection{Morphological Tag Classification}

The joint modeling of segmentation and morphotactic tags %\jason{some of your tags are more than morphotactic, no?  and you rely on that here}
allows us to
use \modelname for  a crude form of morphological analysis:
the task of
\emph{\MTC}, which we define as 
\emph{annotation of a word with its most likely inflectional features}.\footnote{We recognize that this
task is best performed with sentential context (token-based). Integration
with a POS tagger, however, is beyond the scope of this paper.}
To be concrete, our task is to predict the inflectional features
of word type based only on its character sequence and not its sentential
context.
To this end, we take Finnish and Turkish as two examples of languages
that should suit our approach particularly well as both have highly
complex inflectional morphologies.
We use the level 4 tagset and replace all non-inflectional tags with a simple segment tag.
The tagset sizes are listed in \tabref{mtcsize}.\looseness=-1

We use the same experimental setup as in \secref{rootstem}
and compare \modelname to a maximum entropy classifier (MaxEnt), whose
features are character $n$-grams of up to a maximal length of $k$.\footnote{Prefixes and suffixes are explicitly marked.}
The maximum entropy classifier is $L_1$-regularized and its
regularization coefficient as well as the value for $k$ are optimized
on Tune. As a second, stronger
baseline we use a MaxEnt classifier that splits tags into their constituents and concatenates
the features with every constituent as well as the complete tag
(MaxEnt +Split). 
Both of the baselines in \tabref{mtc} are 0$^\text{th}$-order versions of the state-of-the-art CRF-based morphological tagger \software{MarMoT} \cite{muller_efficient_2013} (since our model is type-based), making this a strong baseline.
We report full analysis accuracy and macro $F_1$ on the set of individual inflectional features.

\paragraph{Discussion.}
The results in \tabref{mtc} show that
our proposed method outperforms both baselines on both performance metrics.
%\alex{IMPORTANT: both performance metrics?}
We see gains of over 6\% in accuracy in both languages. This is evidence
that our proposed approach could be successfully integrated into a morphological tagger
to give a stronger character-based signal.

\begin{table}
  \centering
  \begin{small}
  \begin{tabular}{l cc} \toprule
    & Morpheme Tags & Full Word Tags \\
    \midrule
    Finnish & 43  & 172 \\
    Turkish & 50  & 636 \\ \bottomrule
  \end{tabular}
  \end{small}
  \caption{Number of full word and morpheme tags.}
  \tablabel{mtcsize}
  \vspace{-15pt}

\end{table}

\section{Comparison to Finite-State Morphology}
\seclabel{fsmorph}
A morphological finite-state analyzer is customarily a
hand-crafted tool that generates all the possible
morphological readings with their associated features.  
We believe that, for many applications, high-quality
finite-state morphological analysis is superior to \modelname.  
Finite-state morphological analyzers output a small set of linguistically valid analyses of a type, typically with only limited overgeneration.  
However, there are two limitations with finite-state morphological analyzers.  
The first is that significant effort is required to develop the transducers modeling the morphological grammar and creating and updating the lexicon is laborious.  
The second is that it is difficult to use finite-state analyzer to guess
analyses involving roots not covered in the
lexicon.\footnote{While one can in theory put in wildcard
  root states, this does not work in practice due to
  overgeneration.} 
In fact, this is usually solved by viewing it as a different problem, \defn{morphological guessing}, where linguistic knowledge similar to the
features we have presented is used to try to guess POS and
morphological analysis for types with no analysis in the finite-state analyzer.

In contrast, our training procedure learns a probabilistic
transducer, which is a soft version of the type of
hand-engineered grammar that is used in finite-state
analyzers. 
The 1-best labeled morphological
segmentation our model produces offers a simple and clean
representation which could be of use in many downstream applications.
Furthermore, our model unifies analysis and
guessing into a single simple framework. Nevertheless,
finite-state morphologies are still extremely useful, high-precision
tools. 
A primary goal of future work will be  to use \software{Chipmunk} to attempt to induce higher-quality morphological
processing systems.

\section{Conclusion and Future Work}
\seclabel{discussion} 
We have presented labeled morphological segmentation in this paper, a new
approach to morphological processing.  
\LMS unifies three
existing tasks in the literature: unlabeled morphological segmentation, stemming, and \MTC.
%We believe \LMS annotation itself has great potential for  use in
%downstream NLP applications.
Our hierarchy of labeled morphological segmentation tagsets can be used to map the heterogeneous data in six languages we work with to universal representations of different granularities. We plan future creation of gold standard segmentations in more languages using our annotation scheme.

We further presented \modelname a semi-CRF-based model for \LMS that allows for the integration of various linguistic features and consistently out-performs previously presented approaches to unlabeled morphological segmentation.
An important extension of \modelname is embedding it in
a context-sensitive POS tagger. 
Current state-of-the-art models
only employ character level $n$-gram features to model word-internal structure \cite{muller_efficient_2013}.
We have demonstrated that our structured approach outperforms this
baseline. 
We leave this natural extension to future work. 

\section*{Acknowledgments} 
We would like to thank Jason Eisner, Helmut Schmid, Özlem Çetinoğlu as well as the anonymous reviewers for their comments. 
This material is based upon work supported by
a Fulbright fellowship awarded to the first author
by the German--American Fulbright Commission
and the National Science Foundation under Grant
No. 1423276.
The second author is a recipient of the Google Europe Fellowship in Natural Language Processing,
and this research is supported by this Google Fellowship. 
This project has received funding from the European Union's Horizon 2020 research and innovation
programme under grant agreement No 644402 (HimL) and the DFG grant
\emph{Models of Morphosyntax for Statistical Machine Translation}.

\section*{Retrospective}
This version, prepared in April 2024, is a lightly edited version of the original CoNLL 2015 paper.
The Tables are now rendered with the booktabs package and a base case was added to the semi-Markov recursion (\eqref{recursion}).
Finally, a few typos and infelicities in the writing were cleaned up.
All in all, 9 years later, the first author at least still finds the computational analysis of morphology a challenging and unsolved problem.

\bibliographystyle{acl_natbib}
\bibliography{chipmunk}

\end{document}